\documentclass[10pt]{iopart}

\usepackage{url}            
\usepackage{booktabs}       
\usepackage{microtype}      
\usepackage{subcaption}
\usepackage{graphicx}
\usepackage{natbib}

\begin{document}

\providecommand{\newblock}{} 

\title{Repeatable Energy-Efficient Perching for Flapping-Wing Robots Using Soft-Grippers}

\author{Krispin C. V. Broers and Sophie F. Armanini}

\address{Department of Aeronautics, Imperial College London, UK}
\vspace{10pt}
\begin{indented}
\item[]September 2025
\end{indented}

\begin{abstract}
With the emergence of new flapping-wing micro aerial vehicle (FWMAV) designs, a need for extensive and advanced mission capabilities arises. FWMAVs try to adapt and emulate the flight features of birds and flying insects. While current designs already achieve high manoeuvrability, they still almost entirely lack perching and take-off capabilities. These capabilities would enable long-term monitoring and surveillance operations, and more complex and multifaceted missions in cluttered environments.
We present the development and testing of a framework that enables repeatable perching and take-off for small to medium-sized FWMAVs, utilising soft, non-damaging grippers. Thanks to its novel active-passive actuation system, an energy-conserving state can be achieved and indefinitely maintained while the vehicle is perched. This actuation system is inspired by the digital tendon locking mechanism observed in perching birds and allows for high gripping power and minimal energy usage with a low weight penalty.
A prototype of the proposed system weighing under 39\,g was manufactured and extensively tested on a 110\,g flapping-wing robot. Successful free-flight tests demonstrated the full mission cycle of landing, perching and subsequent take-off. The telemetry data recorded during the flights yields extensive insight into the system's behaviour and is a valuable step towards full automation and optimisation of the entire take-off and landing cycle.
\end{abstract}

%
\vspace{2pc}

\vspace{2pc}
\noindent{A Preprint}

%
%
\maketitle
%
\ioptwocol

\section{Introduction}
Bioinspired flapping-wing micro aerial vehicles (FWMAVs) fill in the gap between fixed-wing and multirotor MAVs. They are highly agile and efficient at low flight speeds and small scales, and their extensive flight envelope encompasses hover, glide and cruise flight modes. On top of this, FWMAVs are inherently safe, thanks to their lightweight structures and a propulsion system based on lightweight, often flexible wings rather than hazardous propellers. These features make FWMAVs highly promising for applications in cluttered spaces and delicate environments, especially near humans, flora or fauna. 
Even so, these robots are barely used outside research settings due to their high complexity and limited flight time and payload. Most FWMAVs can only fly for a few minutes \citep{Wu2023} while carrying little or no useful payload and need to be launched and retrieved manually~\citep{Hammad2024}. All this makes complex missions unthinkable without significant improvements.

One approach to enable longer-duration and more complex missions that has been researched extensively in the multirotor community is perching (e.g.~\citep{Culler2012, Doyle2013, Popek2018, Sato2021}), i.e. being able to attach to existing structures in the surroundings. \citet{Culler2012} and \citet{Doyle2013}, for instance, proposed passive, compliant gripping mechanisms for perching on circular or square tubes and demonstrated successful perching of quadrotors in manual guided tests; however, free-flight testing was lacking. The prototype by \citet{Culler2012} weighed 98\,g, while the two-legged mechanism by \citet{Doyle2013} weighed 478\,g.
\citet{Popek2018} demonstrated an active-passive grasping mechanism for multirotor MAVs, achieving, for the first time, automatic free-flight perching capabilities on circular bars, relying solely on onboard sensing and guidance. Their manipulator weighed 372\,g, leading to a system weight of 2.3\,kg \citep{Popek2018}.
\citet{Sato2021} developed a passive gripping mechanism for multirotor UAVs, weighing 640\,g, enabling them to perch on the upper surface of vertical cylindrical objects, which was shown in manually steered flight tests. Over the past few years, numerous studies have investigated different types of perching solutions for multirotors, as reviewed, for instance, in \citet{Roderick2017} and \citet{Meng2022}.

From a perched position, an MAV could perform a useful task, e.g. collect visual data, while consuming only minimal amounts of power. While not actually increasing flight endurance, perching capabilities both allow MAVs to perform longer observation-based tasks and enhance their versatility by enabling them to take off and land on different types of structures.
Perching would, therefore, also be highly useful for FWMAVs, taking them a step closer to real-world applicability. However, perching with FWMAVs is particularly hard due to these vehicles' high complexity, small size, and low weight.

Only recently, a small number of studies have considered the challenge of perching with flapping-wing robots~\citep{Broers2022, Gomez2020, Graule2016, Zufferey2022}.
\citet{Graule2016} achieved perching and take-off utilising an active electro-adhesive pad on a tethered small-insect scale flapping-wing robot. The robot, with its attachment pads, weighed 100\,mg and was able to attach to flat undersides of different materials.
\citet{Gomez2020} developed a shape memory alloy (SMA) activated grasping mechanism for integration onto larger FWMAVs. They demonstrated gripping and static perching capabilities, as well as take-off from a perched position; however, free-flight perching was not achieved.
\citet{Zufferey2022} recently proposed the first successful active-passive perching system, weighing 184\,g for a larger, tailed FWMAV weighing 700\,g in total. It utilises an actuated leg with an attached bistable claw mechanism that passively closes upon contact with a circular branch; resetting the mechanism is realised via a DC motor and completed within 20~seconds. However, a take-off sequence hasn't been implemented. \citep{Zufferey2022}

Perching, therefore, has been achieved under certain constraints in the ultra-small and large size classes, leaving a gap for the highly interesting and widely studied category of small- to medium-sized FWMAVs. This area is particularly challenging as it cannot rely on the different physical conditions that allow for, e.g. adhesive perching at ultra-light sizes, while still being severely weight-constrained to not be able to integrate powerful actuators or actively maintain its perch due to battery limitations. Additionally and importantly for real-life applications, the full free-flight perching manoeuvre cycle of at-will repeatable landing, perching and take-off has seldom been demonstrated. The ultralight flapping-wing robot by \citet{Graule2016}, for instance, represents a successful candidate.

Even in the more extensively studied field of multirotor and fixed-wing aerial robots, only a few were able to demonstrate the required full capabilities. From the previously mentioned examples, \citet{Popek2018} only demonstrated the landing, while \citet{Sato2021} achieved repeated landing and take-off, however, utilising a simpler, solely passive attachment design.

\begin{figure}[htb]
    \centering
    \includegraphics[width=\columnwidth]{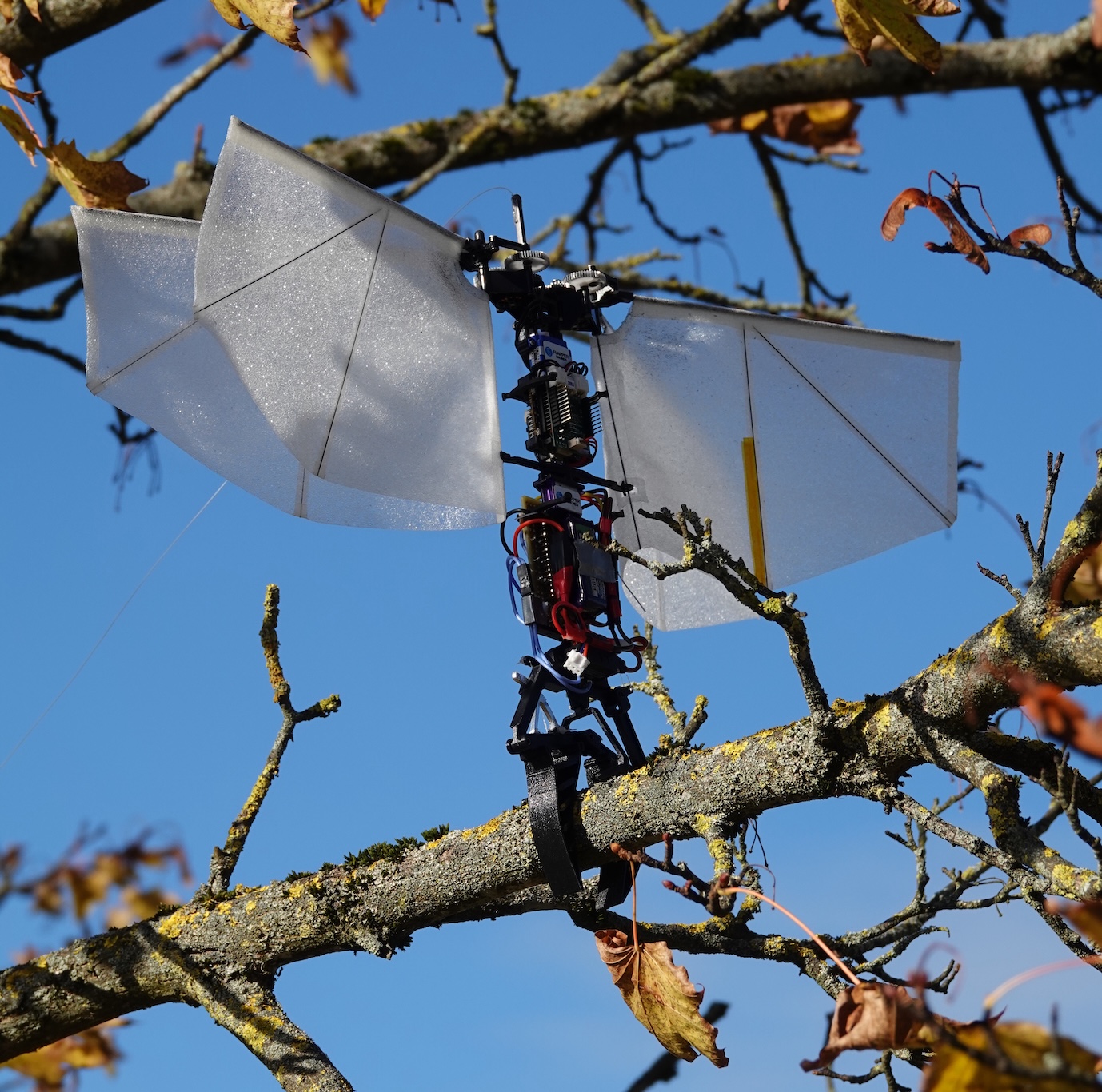}
    \caption{The representative flapping-wing platform, integrated with the proposed perching mechanism, is perched on a tree branch.}
    \label{fig:outdoor}
\end{figure}

In this paper, we present a simple, energy-efficient and non-damaging perching framework for FWMAVs, allowing repeatable perching and take-off cycles during a mission. Building on a soft gripping linkage mechanism developed in our previous research~\citep{Broers2022} and tested on a multirotor platform, we have devised and implemented a system that enables perching functionality on a representative medium-sized (100\,g\,-\,150\,g) tailless FWMAV platform (see figure~\ref{fig:outdoor}).

A central element of our framework is a new, adapted and improved design of the previously introduced soft gripper concept. Key advancements with respect to the original concept include an overhaul of the mechanical design and actuation system, leading to additional capabilities, achieving a full and repeatable perch cycle, as well as weight reductions and performance improvements.
Furthermore, through an integration of the perching mechanism onto a flapping-wing system, extensive tests were conducted, demonstrating our approach's performance and viability, as well as establishing its operation envelope.

Our main contributions towards furthering the capabilities of flapping-wing robots include the following:
\begin{itemize}
    \item Design of a lightweight perching system suitable for small- to medium-sized tailless FWMAVs, which had not been achieved previously. 
    \item Demonstration of the complete sequence of landing, perching, and take-off in a free-flight environment; and validation of the robustness and reliability of the concept through repeated consecutive flight tests. The successful display of a take-off from a perched position, in particular, is frequently neglected in other studies (e.g. in \citet{Popek2018}).
    \item Analysis of data from comprehensive flight and gripping tests of an integrated system to help improve gripper-based perching systems and determine a realistic performance envelope.
\end{itemize} 

\section{Design and Integration}
\begin{figure}[htb]
    \centering
    \includegraphics[width=\columnwidth]{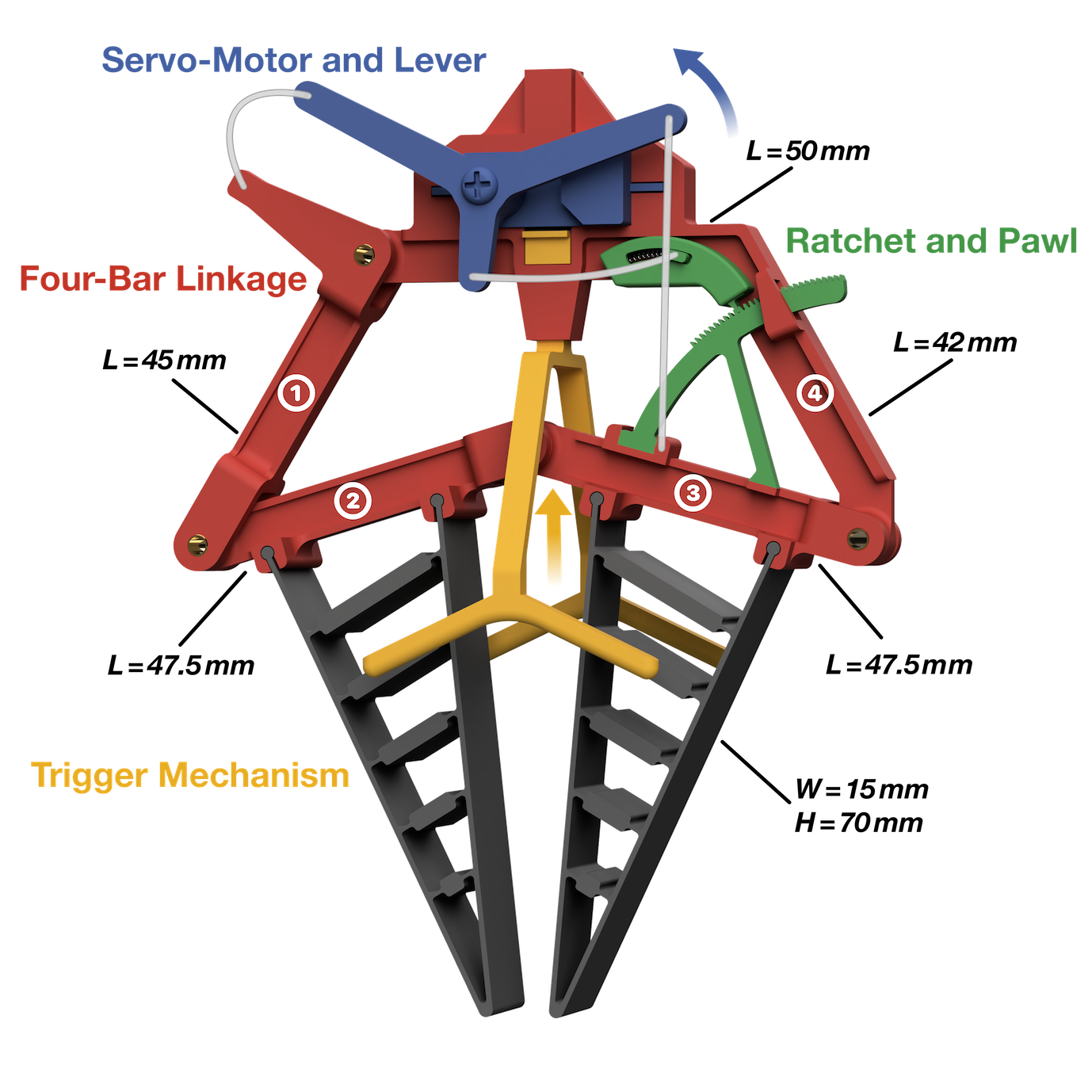}
    \caption{CAD Model showing the gripping mechanism and actuation components of the proposed perching framework.}
    \label{fig:render}
\end{figure}

\begin{figure}[htb]
    \centering
    \includegraphics[width=\columnwidth]{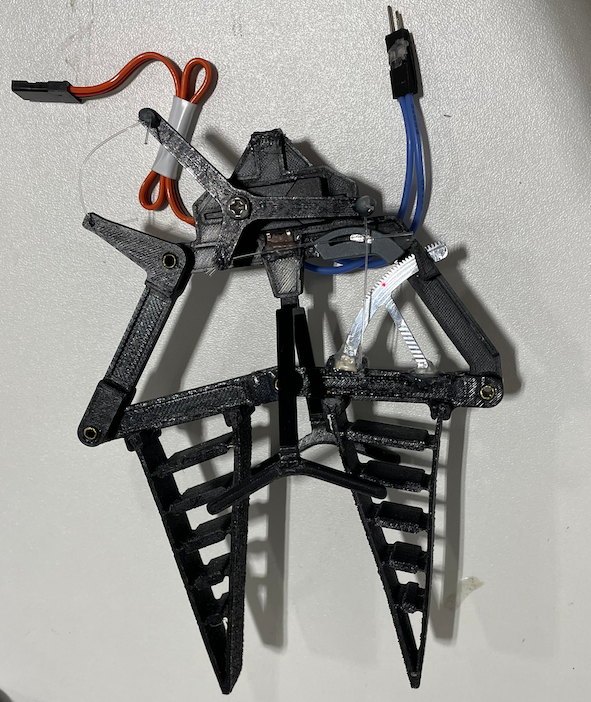}
    \caption{Photo of the assembled gripper prototype.}
    \label{fig:gripper}
\end{figure}

Our perching framework is based on a design concept for a bioinspired gripping mechanism established in \citet{Broers2022}. 
The aforementioned study introduced an initial version of the mechanism, enabling perched landing of an MAV. However, the mechanism did not allow for take-off.
Moreover, while it was ultimately intended for flapping-wing vehicles, the mechanism was only tested on a multirotor and not yet fully adapted for the FWMAV case. 
In this work, we build on the same basic design concept and develop it into a complete perching framework for use on an exemplary FWMAV platform. This included specifically adapting the mechanism design for usage on FWMAVs and fully integrating the mechanism on a FWMAV test platform. Furthermore, the perching capabilities were extended to include take-off and to enable repeatability. Extensive testing was performed to demonstrate and characterise all operation and flight phases of the resulting perching framework.
The resulting design and its components can be seen in figure~\ref{fig:render} with a photo of the produced prototype in figure~\ref{fig:gripper}.

To demonstrate the functionality of our system, we used a commercially available robot as a test platform, i.e. the ``Flapper Nimble+'', produced by \citet{Flapper2024} and based on the DelFly Nimble developed by \citet{Karasek2018}.

The gripping mechanism used in our perching framework was developed using a systematic biomimetic design process proposed by \citet{Fayemi2017} and relies on three main components:
\begin{itemize}
    \item A \textbf{four-bar linkage mechanism}, inspired by the biological jaw structure of parrot fish.
    \item \textbf{Soft and compliant grippers}, based on the Fin Ray effect.
    \item An \textbf{active-passive actuation system} mimicking the digital tendon locking mechanism (DTLM) found in perching birds.
\end{itemize}
The system lends itself to use on FWMAVs thanks to its powerful yet lightweight and efficient design. These features, combined with the soft, adaptable grippers, result in a framework that is versatile and effective, as well as safe for use both indoors and outdoors, including close to humans or in delicate environments. The gripping mechanism is designed to be reliable and robust to aid the hard-to-perform approach manoeuvres of unstable flapping-wing flyers.

A detailed discussion of the gripping mechanism our perching framework is based on can be found in \citet{Broers2022}. Only a concise overview is provided here for completeness, while all the novel elements of our current design are discussed in detail.

\subsection{Mechanism Design}
The gripping mechanism, with its components depicted in figure~\ref{fig:render}, is comprised of an actuated linkage that drives the attached grippers to enclose the desired perching object. This bioinspired linkage mechanism consists of four bars connected through revolute joints (as seen in the numbered linkage components in figure~\ref{fig:render}), with the top bar (4) fixed to the flying platform. The grippers are attached to one of the outer bars (3) and the middle bar (2). The characteristic part of this four-bar linkage design is that it decouples closing and opening forces. For example, applying a closing force at bar (3) results in a scissor-like motion of the linkage, pushing the attached grippers together. On the other hand, applying an opening force at bar (1) straightens the opposing outer (3) and middle (2) bars, making them parallel.

To integrate the gripping mechanism on a generic tailless FWMAV, the top bar (4) is used as a central connection point to the underside of the flying platform. To accommodate various approach paths, a folding symmetry was achieved by introducing a kink in the top bar, mirroring the outer bar (1), which enables the opening functionality.
The bars were 3D-printed from Tough PLA to be lightweight and to provide strong resistance to bending forces, while the linking joints were each made from a polymer sleeve bearing in combination with a hollow brass axle for good friction characteristics.

The soft grippers work according to the Fin Ray effect, which can be observed in fish fins and was first found by Leif Kniese~\citep{Pfaff2011}. The triangular structure of the fin-inspired grippers, containing cross-struts, reacts to a contact force in a compliant manner, bending towards the direction of the impacting force. The basic layout of the grippers was adapted from \citet{Broers2022} and 3D printed from TPU. The geometry consists of a flexible outer shape representing an extruded tall isosceles triangle containing rigid cross-struts hinged through flexible connections to the outer hull. This allows for the desired compliant bending behaviour and soft design, preventing damage to both the perching object and the gripping mechanism while also maximising contact area for various perching object sizes and accommodating irregular surfaces. An attachment to the linkage through a slot system was devised for easy assembly and change of grippers. This leads to increased gripper stability while reducing weight. To optimise the attachment properties of the grippers to the perching object, their inner surface was lined with a thin sandpaper-like anti-slip tape.
The gripper's geometry dictates the range of compatible perching objects. In our study, an opening angle of 35° was chosen, leading to a gripper tip distance of 66\,mm.

\subsection{Actuation Design}
The gripping mechanism is operated in an active-passive manner, similar to how many birds perch. This approach allows for a high achievable gripping power through an active perching mechanism while conserving energy in a perched state through passive engagement. In birds' feet, this is facilitated by the digital tendon locking mechanism, which acts like a ratchet, where the tendon interlocks with a tendon sheath to maintain a perched state~\citep{Quinn1990}.

A servo motor with a power output of 55\,Ncm torque at 7.4\,V was chosen for the active actuation.
For an optimal centre of gravity location, i.e. centred on the Flapper's vertical axis, the servo-motor was integrated into the middle of the top bar.

A lever with three arms was designed for the servo-motor, with a pull cable attached to each of the three levers (see figure~\ref{fig:render}), made from sturdy and lightweight nylon fishing rod:
\begin{itemize}
    \item The right lever connects to one of the outer bars (3) to enable the closing of the gripping mechanism once the servo-motor turns counterclockwise.
    \item The lower and shorter lever connects to the spring-loaded pawl of the ratchet mechanism, which resets the ratchet's locking once the servo-motor turns clockwise.
    \item The left lever connects to the other outer linkage bar (1) and facilitates the opening of the gripping mechanism once the servo-motor turns clockwise.
\end{itemize}

\begin{figure}[htb]
    \centering
    \includegraphics[width=\columnwidth]{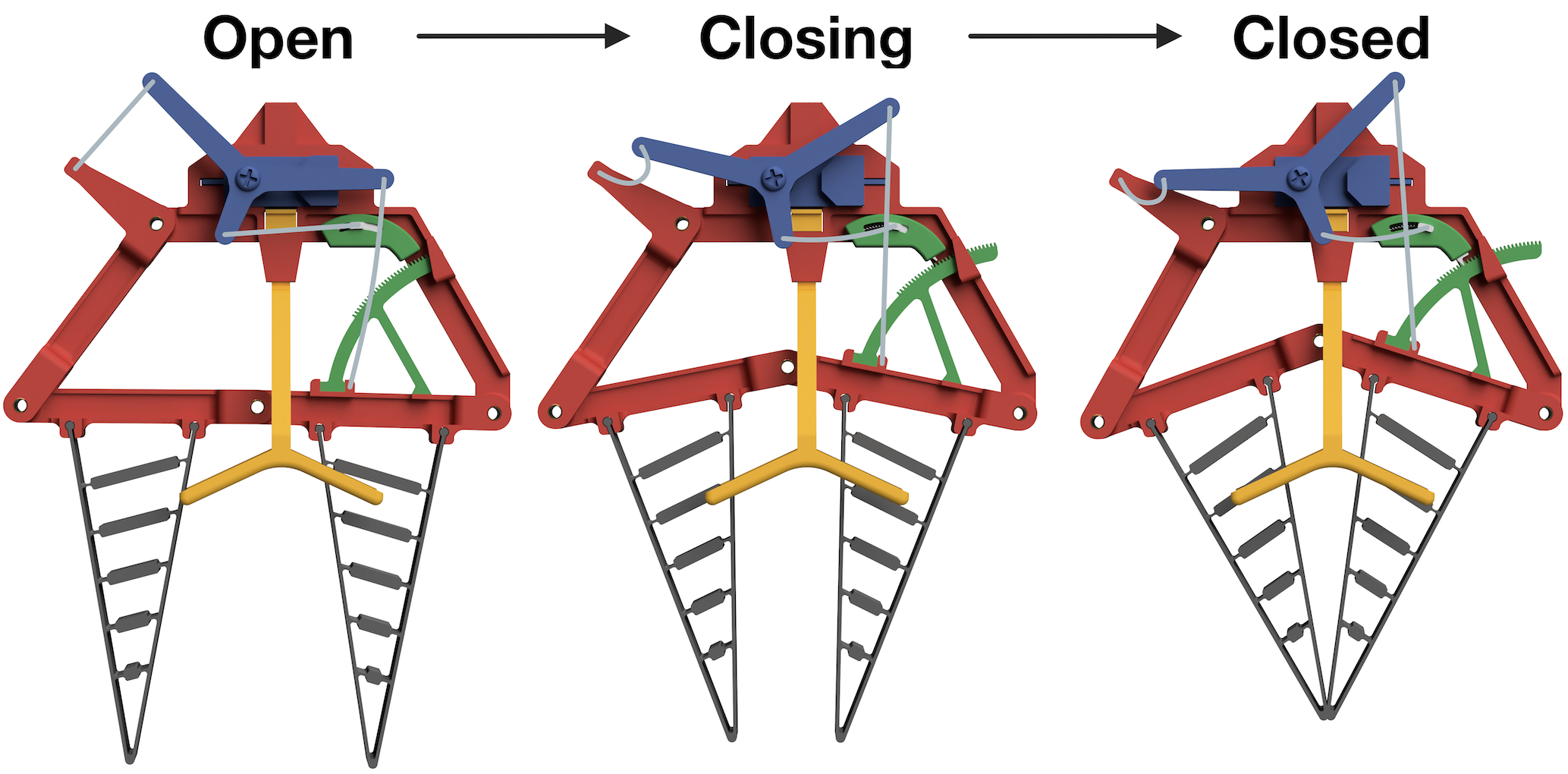}
    \caption{View of the gripping mechanism during the closing sequence.}
    \label{fig:modes}
\end{figure}

Figure~\ref{fig:modes} shows the different positions of the actuation system and grippers during the stages of the closing procedure.

For the passive part, i.e. allowing sustained perching without energy expenditure, a simple ratchet mechanism had been proposed in our previous work~\citep{Broers2022}. However, its rudimentary design could only achieve low precision and, crucially, did not allow for a resetting functionality, thus making an automated take-off from a perched condition impossible.

In the new design, significant changes were introduced to address these limitations, and a new custom spring-loaded ratchet mechanism was developed. The new mechanism consists of a ratchet wheel connected to one of the moving linkage bars and an enclosed spring-loaded pawl that is affixed to the top bar. 
The ratchet wheel was precision CNC-machined from aluminium, enabling good wear resistance and tight tolerances. Its geometry was improved with miniaturised gear teeth, thus maximising gear teeth count for more precise interlocking. Furthermore, the gear tooth flank design was enhanced for the wheel and pawl, enabling a frictionless tightening of the ratchet.
The resulting geometry on the ratchet wheel consisted of 29 evenly spaced teeth over a sector of 37.9° and a root diameter of 64\,mm. The tooth profile is an approximate right triangle, with an angle between the perpendicular flank and the hypotenuse of 18.3°. Additionally, a fillet radius of 0.15\,mm was chosen for the tooth tip and root for manufacturing purposes.

\begin{figure}[htb]
    \centering
    \includegraphics[width=\columnwidth]{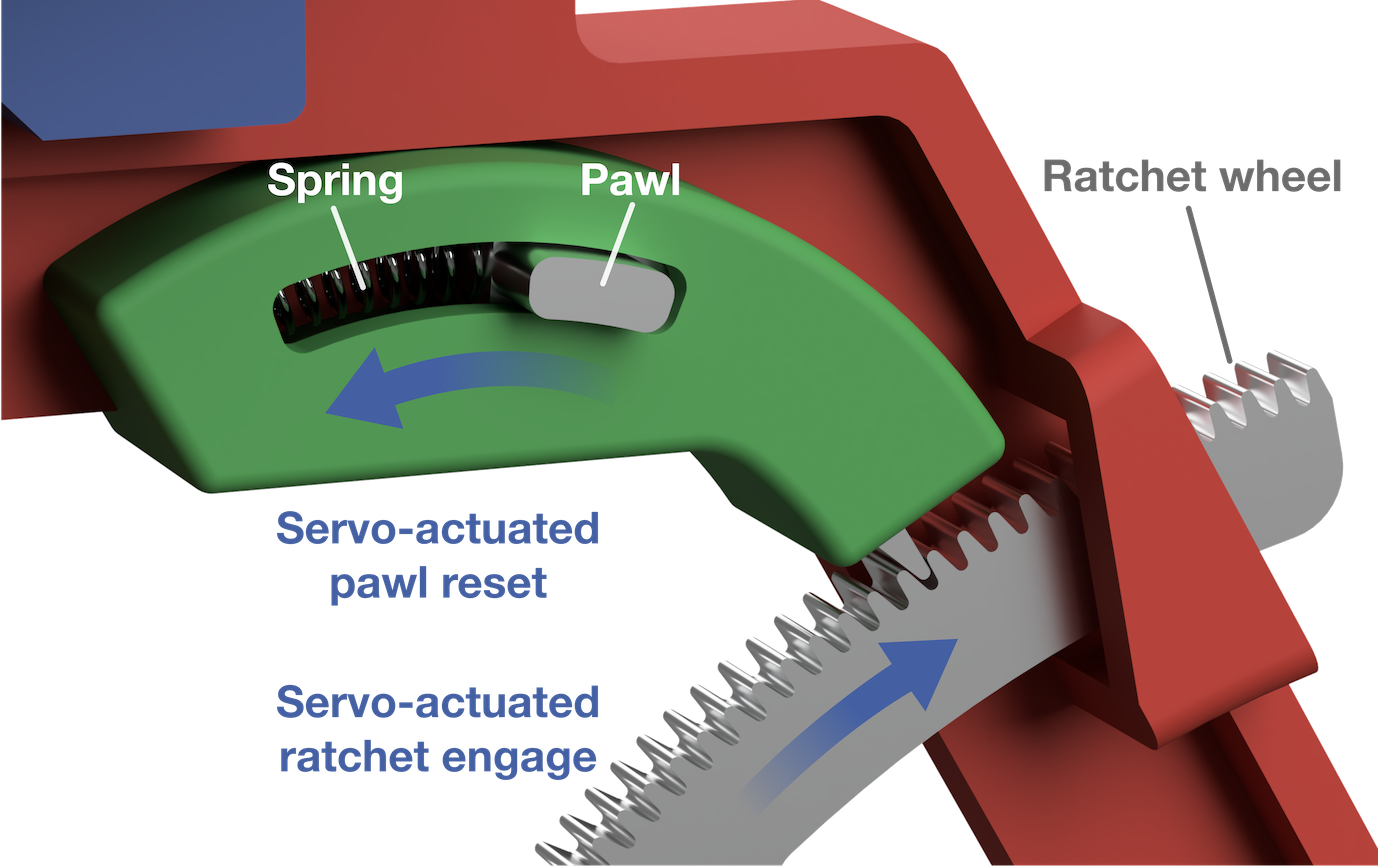}
    \caption{Detailed view of the ratchet's functionality during engaging and resetting.}
    \label{fig:ratchet}
\end{figure}

The curved pawl was likewise CNC-machined from aluminium and included a protruded cable attachment point to enable resetting. It is enclosed in a housing manufactured by precise digital light processing and made out of IGUS Iglidur i3000-PR, a material optimised for sliding parts. Within the housing, a compressed spring constantly pushes the sliding pawl onto the ratchet wheel and therefore allows for the servo-actuated resetting functionality: if the pawl is pulled into the housing, it further compresses the spring and disengages from the ratchet wheel (see figure~\ref{fig:ratchet}).

Through precision manufacturing and the design of the spring-loaded pawl, higher gripping power could be achieved, as nearly all of the servo's actuation power can now be preserved in the passive part of the gripping mechanism.

Actuating the gripping mechanism's closure and thus grasping a perching object is facilitated through a custom trigger system: A trigger fork establishes contact with the perching object and activates a mechanical switch housed in the top bar of the linkage. This basic design was kept from the previous prototype~\citep{Broers2022} but improved upon to account for a flapping-wing platform: Its housing features tighter tolerances and a better attachment to the mechanical switch, allowing for more reliable triggering and higher stability in the perched state. Additionally, the required actuation force of the switch was changed from 2.6\,N to 1.6\,N to accommodate a lighter drone and enable triggering even at lower impact velocities. The selection of the switch model determines the sensitivity of the perching system and, thus, the lower limit of perching contact forces required to initiate the grasping.

\subsection{Assembly and Integration}
In addition to gaining complete landing, perching and take-off functionality and being adapted for usage on FWMAVs, the gripping mechanism has been made lighter compared to the previous proof of concept, weighing 38.8\,g instead of 45\,g. Furthermore, the mechanism features a more compact design, occupying a smaller footprint and containing thinner grippers while exerting larger gripping forces.

This allows for usage on small- to medium-sized FWMAVs that are capable of handling at least this amount of payload. The gripping power mainly constrains the upper size limit of an applicable FWMAV and can, therefore, be defined in different ways. Furthermore, this limit can be easily shifted upward using slightly heavier but increasingly more powerful servo motors.

\begin{figure}[htb]
    \centering
    \includegraphics[width=\columnwidth]{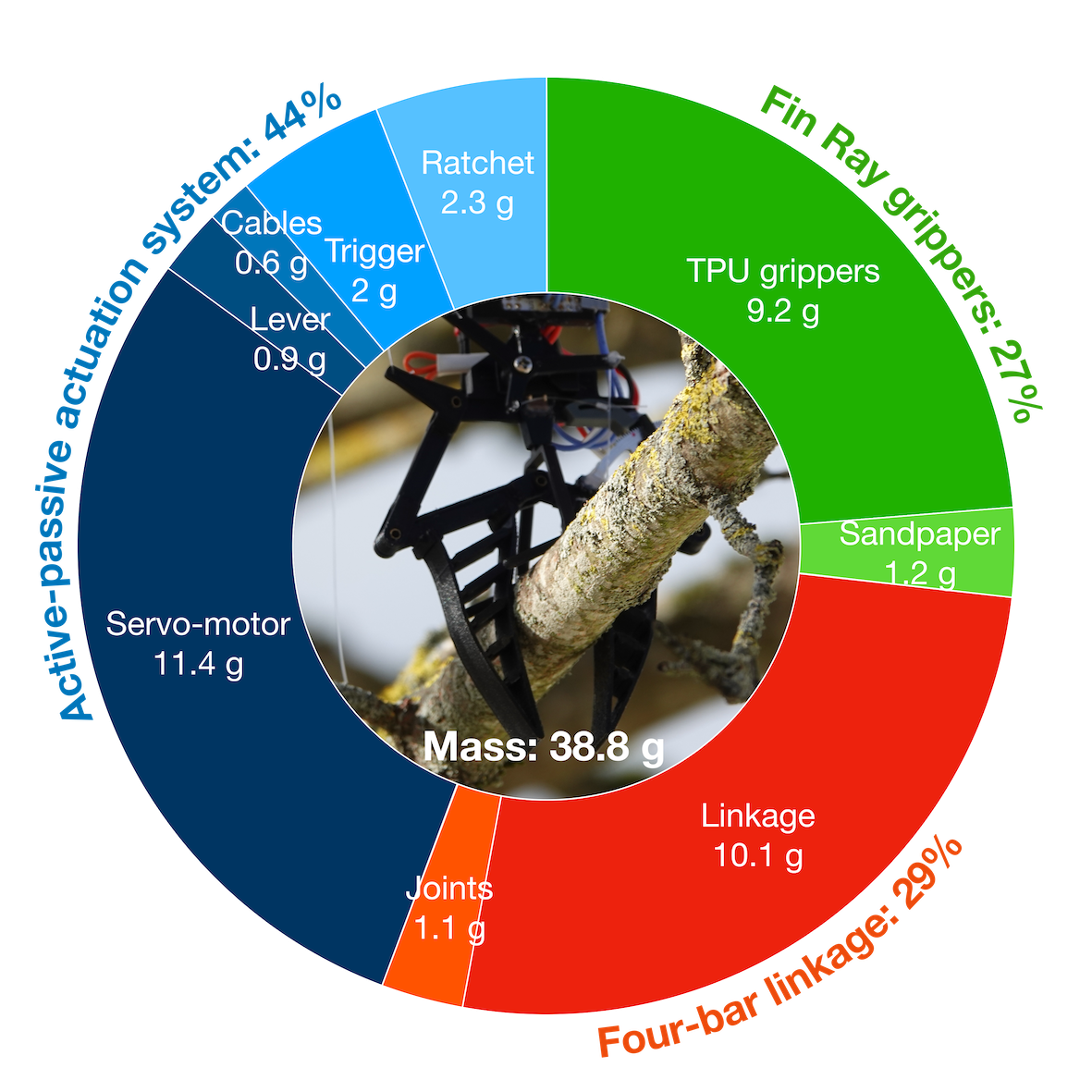}
    \caption{Mass distribution of the gripping mechanism.}
    \label{fig:weightdistr}
\end{figure}

As shown in the weight distribution diagram (see figure~\ref{fig:weightdistr}), the active-passive actuation system takes up the largest part of the total weight at 44\,\%, dominated by the heavy but powerful servo-motor. The ratchet mechanism enables energy-efficient and prolonged perching but only inflicts a negligible weight penalty.
The remaining 56\,\% are almost evenly divided between the mechanical linkage (29\,\%) and the flexible soft grippers (27\,\%).
If this system were to be upscaled to fit larger flapping-wing MAVs, requiring higher gripping strengths, the actuation system would scale favourably and decrease in prominence due to more capable servo-motors.

In order to interface with the active-passive perching system and handle gripping and release states, a power supply and control board had been designed for the first prototype~\citep{Broers2022}. Actuation was controlled by a microcontroller that steered the servo motor and could cut off its power through a MOSFET circuit. For testing purposes and to guarantee modularity for different testing platforms, the circuitry was established as a simple breadboard and not integrated into the platform's onboard electronics.

To reduce weight, the layout of the control board was permanently soldered onto a small and light PCB (weighing 26\,g in total) and a simple integration to any 6\,-\,8.5\,V onboard power supply was established. As this system is not integrated into its potential flying platform's receiver and flight controller, an IR-sensor circuit was included in the board, enabling external reset input using any IR sender (e.g., a TV remote).

To demonstrate the functionality of the perching system and test its capabilities and suitability for FWMAV, a representative commercially available drone was chosen for testing, i.e. the ``Flapper Nimble+'' (see figure~\ref{fig:outdoor}) produced by \citet{Flapper2024}. The Flapper is a lightweight and tailless flapping wing MAV with hover capabilities, enabling a wide variety of flight manoeuvres. Its four wings, consisting of two pairs, are arranged in an X-configuration with a wingspan of 490\,mm. The flapping motion for each wing pair is created by a brushless DC-motor-powered gear train and linkage, resulting in a flapping frequency of 20\,Hz at full throttle. Manoeuvres are achieved through wing geometry and wing motion alterations, which are enabled by two servo motors and two independent brushless DC motors. As the tailless configuration makes the Nimble+ inherently unstable, a Crazyflie Bolt 1.1 flight controller is used. The complete system is powered by a single 300\,mAh 2S LiPo battery, enabling a flight time of more than five minutes hovering with a payload of 25\,g~\citep{Flapper2024}.

For integration into the proposed system, the Flapper Nimble+ was used in its bare configuration (i.e.~without the detachable landing gear, body shells, and head shell) to minimise take-off weight. In addition, the control board was affixed to its carbon fibre spine, which was opposite the battery. The gripping mechanism was then connected to the bottom of the spine by inserting the squared spine into the counterpart fixture on the top bar of the linkage.
An alternative approach would be to attach the gripper system on top of the FWMAV's spine, allowing for perching from below, with the robot hanging on to the perching object. This would result in a more stable perch and would be an interesting approach to test in future work. However, positioning the gripper at the top would shift the center of gravity upwards, introducing further instability into the already unstable flying platform.

The power supply of the Flapper was connected to the control board and the drone via a split connector. Lastly, the gripping mechanism was attached to the board with small detachable connectors, allowing for quick exchange during maintenance or a change of component position.

After integration, this puts the combined system's weight, including the flapper drone (89\,g), battery (18\,g), gripping mechanism (38.8\,g), and control board (26\,g), at roughly 172\,g of take-off mass. Note that this exceeds the recommended take-off weight, primarily due to the external control board; however, for our testing purposes, the resulting shortened flight time was acceptable.

\section{Testing and Results}
Three different stages of tests were conducted after the functional perching mechanism was completed and integrated onto the FWMAV. Given that the performance of different gripper configurations was already tested in our previous work, the current tests could directly concentrate on the performance of the full, integrated system. 

The first stage consisted of static tests, in which the system's gripping performance was assessed against disturbance forces and moments of different magnitudes and orientations.
To provide a complete and comprehensive evaluation of the system, we tested all three disturbance force directions, as well as the resulting disturbing moments that act on the integrated system. 

Additionally, we also accounted for a special ``upside-down'' perched case and different perching-object inclinations to get a fuller understanding of the robustness of the system while perched. 
Considering different inclinations provides an overview of the set of perching conditions that can successfully be handled by our mechanism and is essential because, in a real-world setting, branches, pipes or railings will often not be perfectly aligned with either the ground or the robot.
The ``upside-down'' perching case is important because our perching mechanism, like many others, is attached to the lower end of the robot, while the CG of our robot is relatively high up: therefore the perched robot is particularly susceptible to transversal forces, which may make it lose its balance and topple over. However, if the gripping force is such that the robot remains in a hanging, bat-like position, rather than drop off the perch, the rolling over motion becomes less problematic -- given sufficient space, the robot is expected to be able to take off from an upside-down position. 

During the second stage, free-fall tests were conducted to investigate the system's capabilities during different approach scenarios through variations in velocity and approach angle.

Lastly, free-flight tests under realistic conditions were carried out, demonstrating the complete perching cycle of landing, holding on, and taking off.

\subsection{Static Performance Testing}\label{sec:statictests}
In the first testing stage, the performance of the perching system was evaluated through manual placement of the integrated drone on a variety of perching objects. During these tests, the gripping strength and resilience against disturbing forces in all directions were quantitatively measured using a force gauge.

\begin{figure}[htb]
    \centering
    \includegraphics[width=\columnwidth]{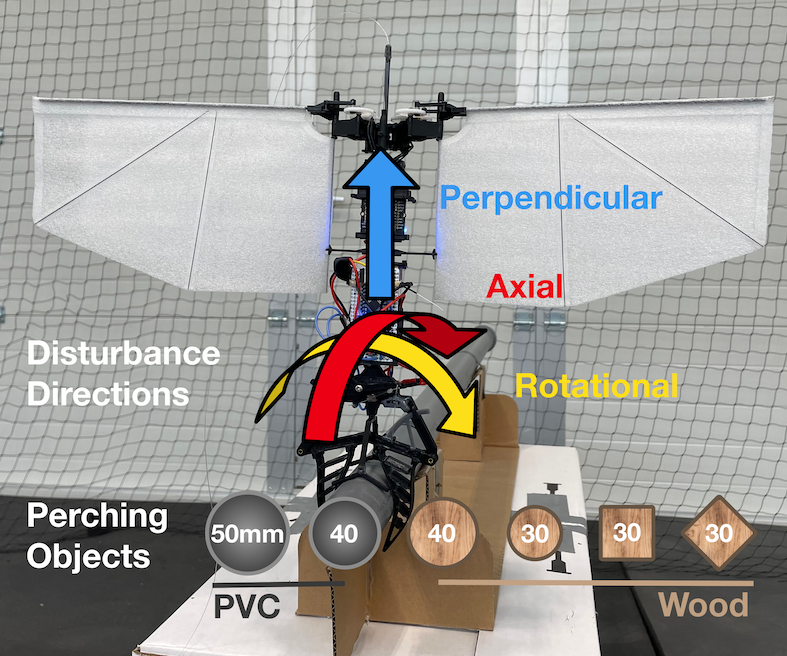}
    \caption{Test setup for the static performance tests showing the integrated system perched on a pipe placed on a fixture. Directions of measured disturbance moments and forces are shown. Additionally, the variety of perching objects tested is shown to scale.}
    \label{fig:statictest}
\end{figure}

The testing setup, which can be seen in figure~\ref{fig:statictest}, consisted of a fixture in which different perching objects were placed and oriented at a desired inclination angle (see also figure~\ref{fig:inclination}). To repeatably and reliably assess perching performance, a variety of perches were used for these tests, i.e.: 
two circular PVC pipes, with diameters of 50\,mm and 40\,mm respectively, two circular wooden bars, with diameters of 40\,mm and 30\,mm, respectively, and a square wooden bar, with a cross-section of 30\,mm by 30\,mm, which was used both in a flat square orientation and a diamond orientation (see figure~\ref{fig:statictest}).
These objects were selected to represent common indoor and outdoor landing structures, like tree branches, railings, and struts.

A force gauge was connected to fixtures at specific positions of the drone to measure performance. Four different measuring directions were used to represent disturbance forces or torques for each degree of freedom (see coloured arrows in figure~\ref{fig:statictest}). The first measurement was taken horizontally in a perpendicular manner to the perch at a fixed distance from the pivot point, in order to measure the rotational momentum around the perch (yellow arrow). The second measurement was taken axially to the perch, again at a fixed distance to the pivot point, in order to measure the axial moment of possible disturbances (red arrow). In this way, the peak force was measured until the system's slippage point. The remaining two measurements were taken vertically, perpendicularly to the perch, once measuring upwards with the system sitting atop the perching object (blue arrow) and once downwards, with the system ``hanging'' below the perch. The results of these perpendicular measurements represent the maximal force required to loosen the grip from the perch, a good indicator of general gripper strength.

Each of the four measurement types was repeated three times for each of the six different perch configurations. Each test consisted of manually bringing down the system from a low height onto the perch, thus automatically triggering the gripping process, and then performing all measurements on the freely perching drone with the servo motor shut off.

\begin{figure}[htb]
    \centering
    \includegraphics[width=\columnwidth]{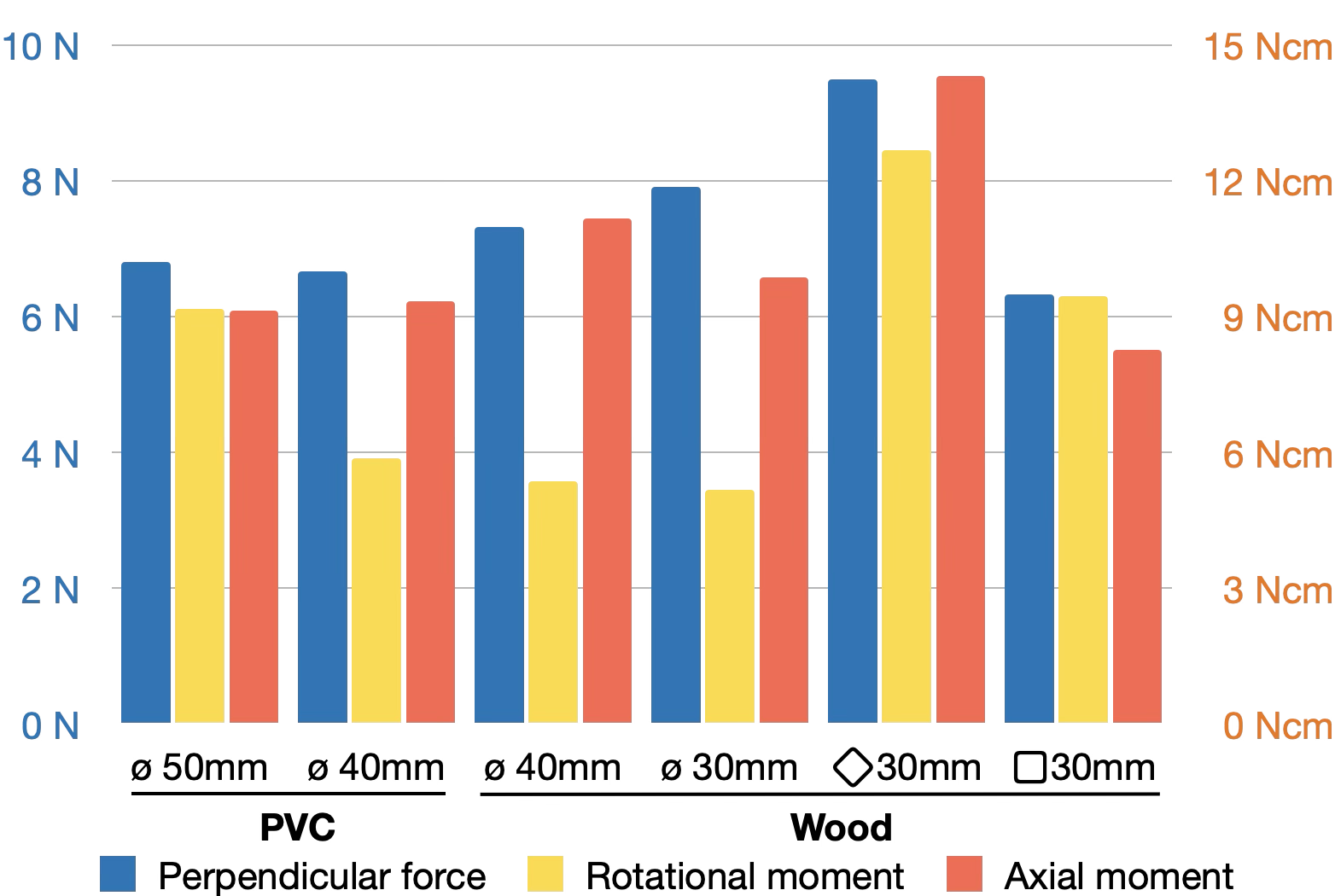}
    \caption{Overview of the perching performance of the system on a variety of perching objects during static performance testing.}
    \label{fig:perchperformance}
\end{figure}

The results of these tests are shown in figure~\ref{fig:perchperformance}, which presents the resulting forces and moments in each of the three main measurement directions clarified in figure~\ref{fig:statictest} (with the perpendicular down direction excluded here for clarity), for each of the considered perching objects.

The perpendicular forces needed to pull the system from its perch are consistently above 6\,N, with an arithmetic mean of 7.4\,N across all perching objects. The best-performing scenario was the diamond-oriented square wooden bar, which required 9.5\,N of force. This can be explained by the better gripper performance when conforming to irregular shapes. Here, especially, the contact point of the object is in a good position, not too far out to the gripper tips, but acting on the middle portion, allowing the gripper to close further compared to a square-oriented object.
The gripper strengths show a mean improvement of 42\,\% more force required in the upwards perpendicular direction, normalised with the system weight compared to the results obtained in previous work for a similar design concept~\citep{Broers2022}. This gain can be attributed to the generally optimised design and especially the overhauled ratcheting functionality.
The rotational disturbance moment required for continuous displacement around the perch shows an increase with larger contact area and irregular shapes, with the smaller diameters of perching objects performing the worst as they move further from the primary gripper size design point. The best-performing case is again the square wooden bar in a diamond orientation, which can withstand a 12.7\,Ncm moment of disturbance.
The mean gripper strength, measured through its resistance to a rotational disturbance moment, increased by 63\,\% compared to the previous design~\citep{Broers2022}.
Lastly, the axial disturbance moment was largely consistent across the different perching objects, with a mean of 10.3\,Ncm of disturbance required for continuous displacement in the axial direction. Consistent with the other disturbance measurements, the diamond-oriented square wooden bar performed best, with a moment of 14.3\,Ncm required for displacement.

For further testing, the ø\,40\,mm circular wooden bar was chosen, as it consistently led to the best perching performance. While better results were achieved with the diamond-oriented square bar, these were highly dependent on the bar's orientation, as can be seen in the results for the flat square orientation. Consequently, the round bar was deemed more suitable for systematic testing. A cylindrical shape is also more generic and more closely approximates typical perching structures in real-life scenarios. Nonetheless, the results obtained on the diamond-shaped cross-section suggest that irregular shapes, such as natural tree branches, for example, can yield better results than the precisely cut round testing object chosen here. This not only suggests that our proposed mechanism will be particularly suitable for outdoor applications, but also implies that our subsequent test results will be somewhat conservative. 

Thus, in addition to the basic measurements discussed, further tests were conducted on the ø\,40\,mm circular wooden bar to fully explore the feasibility envelope of our mechanism. Specifically, the previous tests were repeated at three further inclinations of the perch axis: 5°, 10°, and 12.5°. The highest inclination of 12.5° was selected following preliminary tests that indicated this to be close to the maximum inclination that could still successfully be handled.
For these tests, the fixture of the bar was lifted vertically on one side, resulting in a defined inclination of the perch axis compared to its horizontal state (see also figure~\ref{fig:inclination} for orientation reference).

\begin{figure}[htb]
    \centering
    \includegraphics[width=\columnwidth]{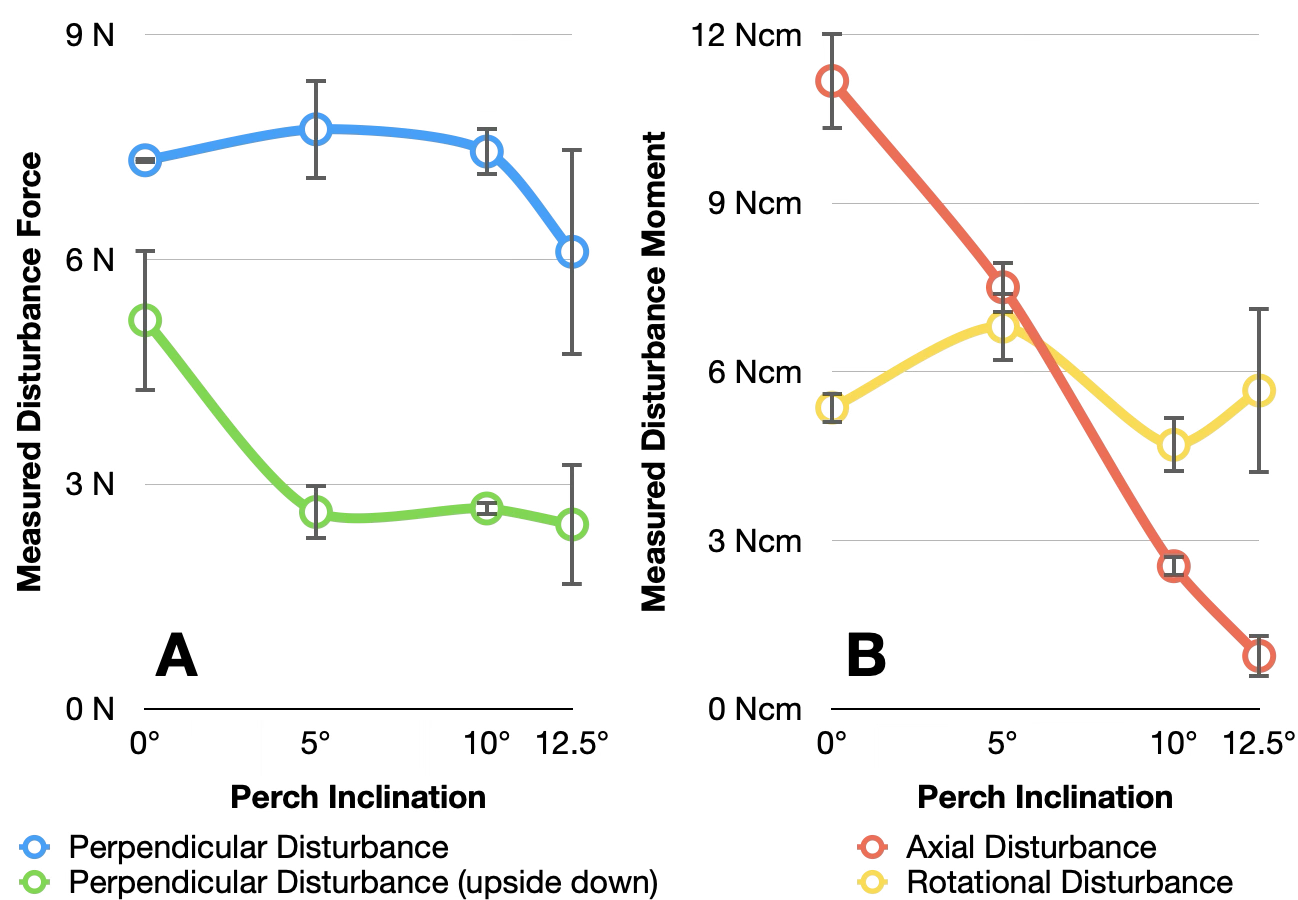}
    \caption{Results from static perching performance testing on a ø\,40\,mm wooden bar for a range of possible inclinations.}
    \label{fig:perchinclination}
\end{figure}

Results of these further tests can be seen in figure~\ref{fig:perchinclination}:
In panel A, the perpendicular disturbance force for both the regular and the upside-down direction is plotted against the different perch inclinations. For the upwards perpendicular force, the results are fairly similar up to 10° of inclination, followed by a significant drop in performance at the highest inclination of 12.5°. For the downwards perpendicular direction, performance drops off at an inclination of 5°, where it remains stable even for higher inclinations. Of note for the upside-down case is that even for the highest tested inclination, roughly 300\,g of weight are still supported. This indicates that even in a scenario where the robot loses its balance and reverses, it will not drop off the perch, acting as a useful fallback.

In panel B of figure~\ref{fig:perchinclination}, the rotational and axial disturbance moments are plotted against the perch inclinations. The rotational moment required for displacement increases somewhat for 5° of inclination, compared to the non-inclined case, but stays fairly consistent across all tested inclinations overall. However, a sharp linear decline in the measured moment is seen for the axial disturbance moment, nearing zero at the highest inclination setting. This is in line with expectations, as the highest inclination attainable is defined by the system's robustness against axial moment, which increases with perch inclination as the distance between the pivot point of the system and its centre of gravity increases.

Based on these results, attachment strength measured through the perpendicular disturbance force is well-performing at inclinations up to 10°. This would enable the system to find a large variety of perching opportunities on natural structures such as inclined tree branches and man-made struts.
However, the resistance against axial disturbance forces after successful perching attempts is severely diminished at these higher inclinations. Therefore, if robustness against axial disturbances is required, e.g. in windy outdoor scenarios, perching objects of lower inclination will be strongly preferred. To further increase this resistance, especially at higher inclinations, an optimisation of the gripper surface material would be required to provide better adhesion to the perching object.
The full feasibility envelope is further explored in testing in the following section.

\subsection{Free-Fall Testing}
In the second testing stage, free-fall tests were conducted to assess perching behaviour at different approach velocities and angles. The setup consisted of the aforementioned perching object fixture with the ø\,40\,mm circular wooden bar installed. During these tests, the Flapper drone was held directly above the perch at different heights and manually released, thus falling onto the perch at different impact velocities. In order to gather information on the approach, impact and gripping behaviour, two flight data recording systems were used: a Vicon motion capture system and the internal IMU (Bosch BMI088) in the drone's flight controller (Bitcraze Crazyflie Bolt 1.1). Both were enabled simultaneously in the first test run; however, the internal recording of IMU data was solely used via a microSD card later during more extensive tests, as it proved to reliably yield similar results.

These data recordings provide insight on the behaviour of the system during and after impact, especially on the exact timings of the gripping operation and the elasticity and deformation of the trigger components. Additionally, they allowed for an accurate reconstruction of the impact velocity.

For each test, the drone was dropped from increasing heights, and the behaviour of the gripping mechanism and the whole system was noted. Overall, 90 fall tests were conducted in order to find the lowest possible approach velocity leading to successful perching, and, through continuous fall height increases, the highest feasible approach velocity. Due to the manual release and uncontrollable changes in external conditions, it was impossible to repeat any two tests during these fall tests. However, many tests were performed at similar speeds with an overall range of tested velocities for a fast approach of 0.77\,-\,1.29\,m/s.

\begin{figure}[htb]
    \centering
    \includegraphics[width=\columnwidth]{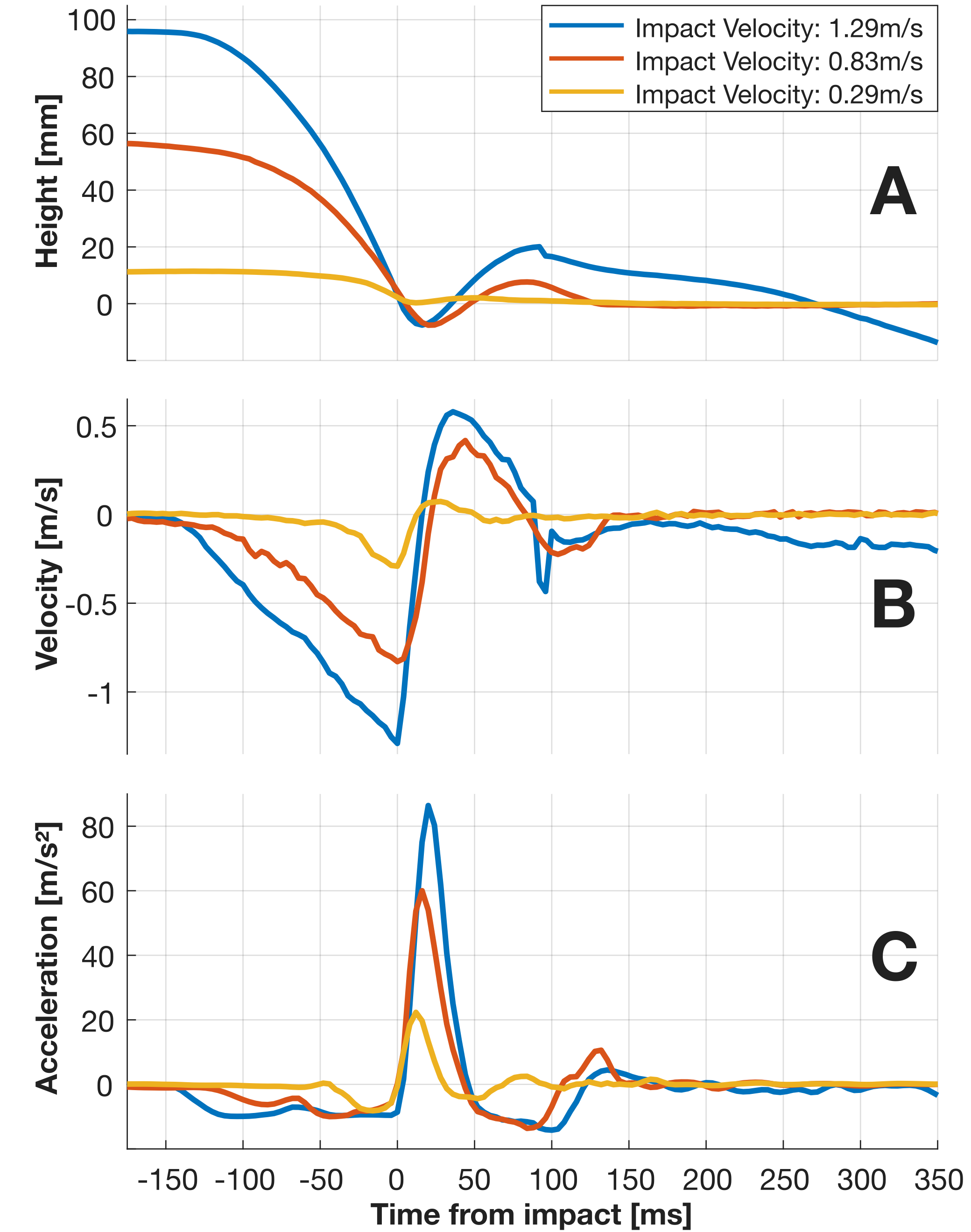}
    \caption{Overview of data captured from three free-fall tests at different impact velocities: firstly, a successful perching attempt at low impact speed; secondly, a successful attempt at high impact speed; and lastly, a failed attempt at too high impact speed. Panel A represents the positional data from the Vicon motion capture system, with panel B showing the calculated velocity from that positional data. Lastly, panel C contains the IMU-supplied acceleration data.}
    \label{fig:freefall}
\end{figure}

The fall test data from the two sources was then post-processed to obtain the relevant information:
\begin{itemize}
    \item The Vicon motion capture system provided positional data at 250\,Hz in a Cartesian coordinate system with a set zero-point. For our tests, primarily the Z-position, i.e. height, was of interest. Furthermore, the velocity of the system was determined by calculating the derivative of the positional data. The impact velocity for the free-fall tests is represented by the maximum gradient value recorded.
    \item The flight controller's IMU provided acceleration data at 250\,Hz to an onboard microSD card. Again, the Z-acceleration, i.e. vertical acceleration, was of primary interest. With the identification of the two relevant zero-crossings before the peak of maximum acceleration (clearly seen in figure~\ref{fig:freefall}.C), the velocity at impact was calculated via integration.
\end{itemize}

The first test run consisted of over 26 recordings at different approach speeds, utilising both data sources simultaneously. These revealed that the Vicon and IMU data yielded similar and plausible results for the calculated impact velocities, with a mean velocity discrepancy of 6\,\% and a standard deviation of 5\,\%. In Extension~\ref{Extensions}.1, a table of these data recordings can be accessed. In figure~\ref{fig:freefall}, three distinct approaches are shown from the free-fall testing.
The figure contains Vicon-measured height data (panel A), Z-velocity calculated from the Vicon position data (panel B), and IMU-measured Z-acceleration data (panel C).
The Vicon and IMU datasets are synchronised from the time of impact, which was set at the data point corresponding to the highest approach velocity.
The three examples showcase three different impact velocities, ranging from  0.29\,m/s to 1.29\,m/s. While the lower and middle velocity still lead to a successful perch, the highest velocity case is an example of an unsuccessful perching attempt.

Of note in panel A is the bounce-back after impact, which is mainly caused by the elastic trigger fork design, increasing in height with higher impact velocities. Likewise, in panel A, the unsuccessful attempt (blue line) can be easily identified, as the re-attachment to the perch does not happen, i.e. rather than stabilising at zero (perch-level), the height continues to drop after impact. Similar effects can also be seen in panel C, where the bounce-back effect is clearly visible in the positive Z-acceleration peaks.
Higher impact velocities, therefore, lead to a quicker and larger recoil of the system from the perch, at one point exceeding the gripper reaction time or, rather, leaving the designed closing area too early.

\begin{figure}[htb]
    \centering
    \begin{subfigure}{\linewidth}
        \centering
        \includegraphics[width=\linewidth]{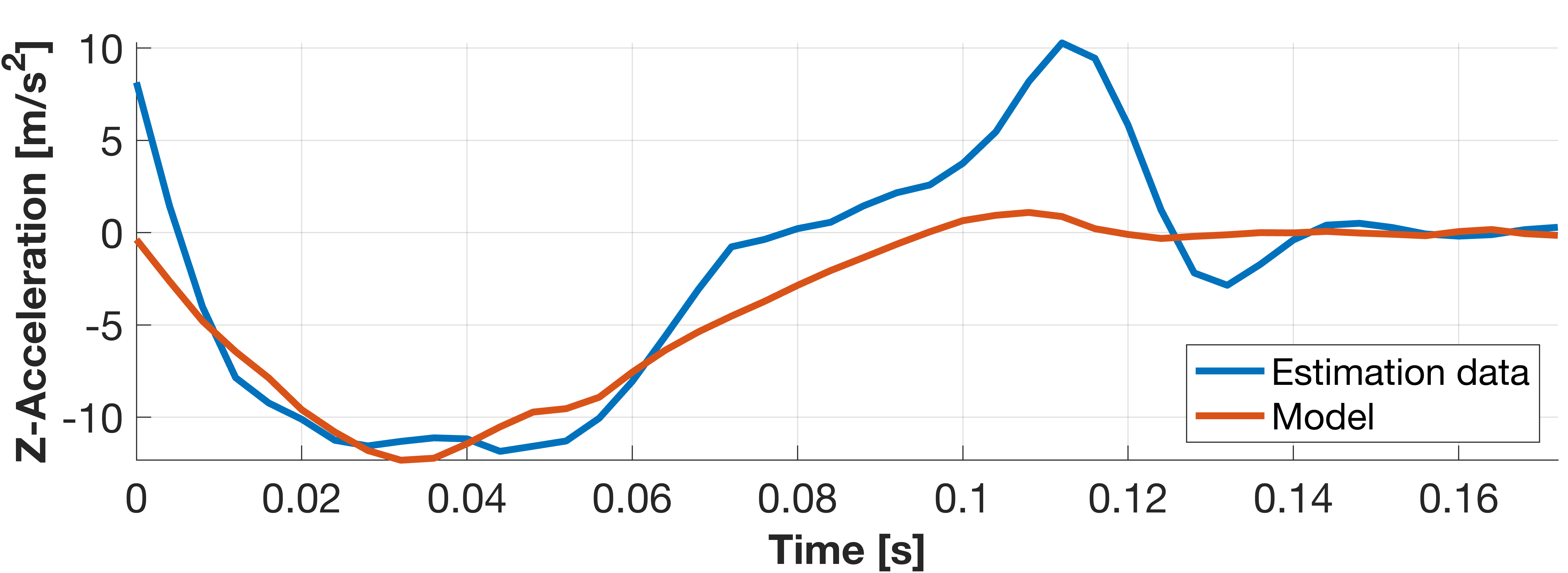}
        \caption{Model estimation using data from a free fall test with 1.0\,m/s impact velocity. The Pearson correlation coefficient is 0.903.}
        \label{fig:estimation}
    \end{subfigure}

    \vspace{1em}

    \begin{subfigure}{\linewidth}
        \centering
        \includegraphics[width=\linewidth]{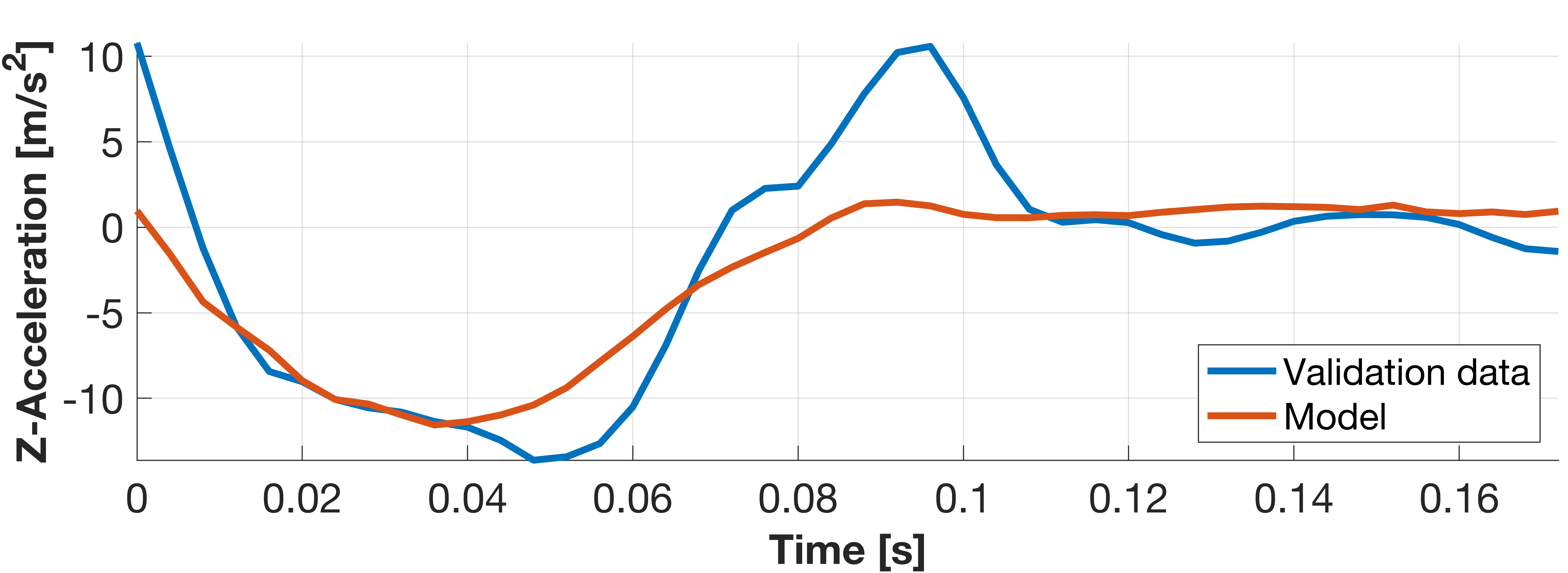}
        \caption{Model validation using data from a free fall test with 0.83\,m/s impact velocity. The Pearson correlation coefficient is 0.873.}
        \label{fig:validation}
    \end{subfigure}

    \caption{Visualisation of the robot's Z-acceleration for a time period shortly after impact until relative standstill, used for estimating and validating the 1-DoF mass-spring-damper model. The estimated coefficients from the linear regression model of the spring-mass-damper are the spring stiffness k = 235.8\,N/m and the damping coefficient c = 2.44\,Ns/m.}
    \label{fig:model}
\end{figure}

In order to quantify the properties of the system after impact, we modelled it as a 1-DoF mass-spring-damper system and estimated the spring stiffness \textit{k} and damping coefficient \textit{c} using a linear regression with a dataset from the free fall tests (see figure~\ref{fig:estimation}). This resulted in an estimated spring stiffness k = 235.8\,N/m with a standard error of 22.2\,N/m, and an estimated damping coefficient c = 2.44\,Ns/m with a standard error of 0.5\,Ns/m. We validated this model with another dataset of a fall test with lower impact velocity and achieved similar accuracy (see figure~\ref{fig:validation}).

\begin{figure}[htb]
    \centering
    \includegraphics[width=\columnwidth]{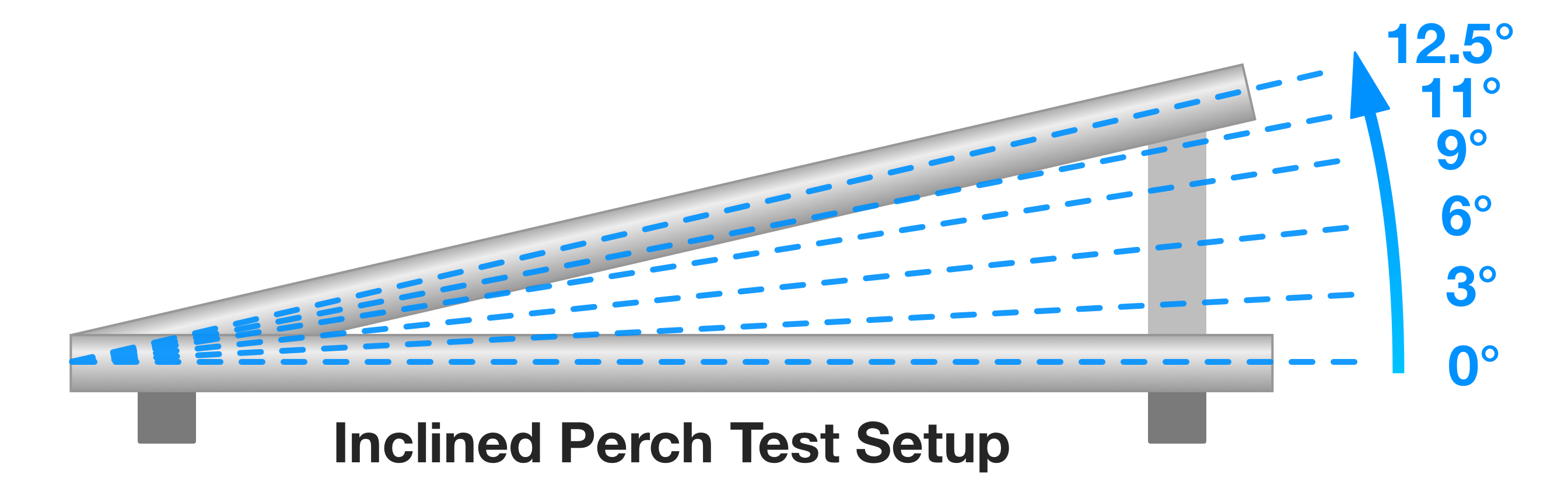}
    \caption{Schematic showing the side of the perch test fixture and the inclinations chosen for the free-fall tests.}
    \label{fig:inclination}
\end{figure}

As the recording of the Vicon data adds complexity to the setup and adds weight to the robot, due to the markers that need to be attached to it, all further 64 fall tests were conducted using only the on-board IMU data recording. A table containing the calculated velocities from these tests can be accessed in Extension~\ref{Extensions}.2. The aim of these tests was to determine the maximum feasible impact velocity through gradual variations of the drop height.
These tests were conducted for five different perch inclinations, i.e. 0°, 3°, 6°, 9° and 11° (see visualisation in figure~\ref{fig:inclination}), and the maximum successful impact velocity was obtained for each of these cases. 

\begin{figure}[htb]
    \centering
    \includegraphics[width=\columnwidth]{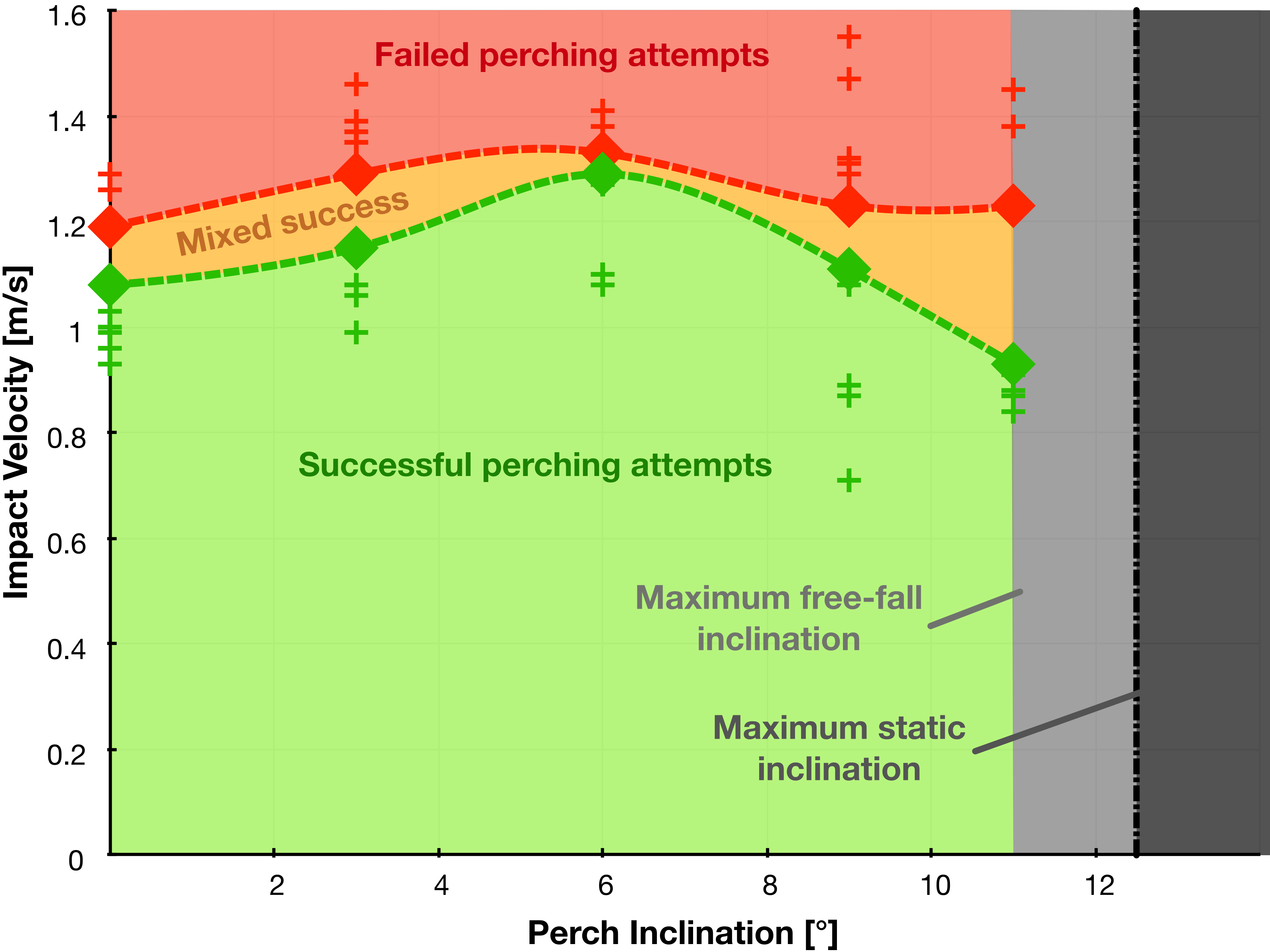}
    \caption{Sufficiency subspace showing feasible perching impact velocities for different perch inclinations, resulting from free-fall tests, with data captured from the onboard IMU.}
    \label{fig:subspace}
\end{figure}

At least five data points each, for successful attempts and for unsuccessful attempts, were collected per inclination setting. Results of these tests are displayed in figure~\ref{fig:subspace} in the form of an approach sufficiency subspace, where the different categories of perching attempt outcomes (failed/successful/mixed) are shown in a plot of impact velocity over perch inclination. The green area in the diagram represents a 100\,\% success rate of perching attempts, while the red area represents predominantly failed perching attempts. The green and red crosses represent individual data points that fall within each category. The orange area in between represents a mixed success rate of perching attempts.
Along the x-axis, the grey band shows the maximum inclination that was successfully tested in free-fall tests (just over 11°), while the dashed black line represents the maximum inclination achievable in the static tests (cf. section~\ref{sec:statictests}), i.e. 12.5°. This discrepancy between both inclination limits is due to disturbances introduced during the approach and the recoil forces acting on impact.

The maximum inclination is where the gripping strength is overcome by the axial moment induced by the system's shifted centre of gravity position. The black band represents inclinations where successful perching could not be obtained in any of the tests. 

In general, the velocity at which the failure rate becomes dominant stays fairly constant, at around 1.3\,m/s, across different inclinations. However, the highest feasible impact velocity that guarantees a successful attempt declines sharply at higher inclinations and has a maximum for the 6° inclination at 1.3\,m/s. This maximum could possibly be explained by a small offset due to an unintended and imperfect centre of gravity location of the integrated system, where a slight inclination might lead to a better balance on the perching object, with optimum stability at around 6°. Alternatively, this could be due to a slightly larger and asymmetric gripper contact surface area due to the change in the overlapping area between the grippers and the inclined perching object. For even larger inclinations, the rising axial momentum of the system becomes the driving factor and diminishes these gains again.
Overall, the highest impact velocity leading to 100\,\% successful perching attempts, up to and including 9° of inclination, is 1.1\,m/s.

In these tests, the maximum feasible impact velocity is strongly dependent on the trigger design, especially its ability to absorb impacts through dampening and elasticity, which both regulate the extent to which the system recoils from the impact. This parameter could potentially be tuned for a different platform and desired approach speed, with the limiting factors being the additional weight of a more complex trigger design (e.g. with dampening capabilities) and the reaction time of the gripping system. The minimum viable impact velocity, on the other hand, is dependent on the weight of the whole system and the sensitivity of the trigger design. The latter is again a design parameter that can be fine-tuned depending on the specific system and mission requirements. 

\subsection{Flight Testing}
\begin{figure*}[t]
    \centering
    \includegraphics[width=\linewidth]{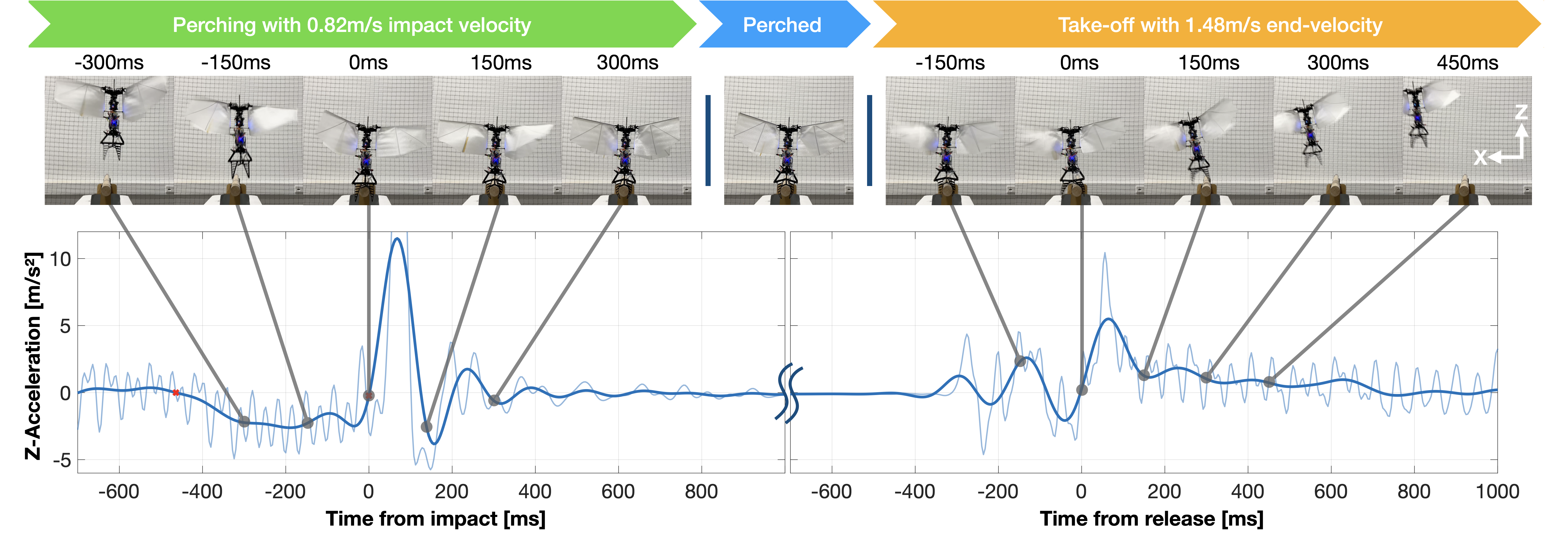}
    \caption{Example of the full perching cycle of the system obtained during free-flight tests. Acceleration data from the IMU, excluding gravity, is plotted against time, with accompanying time frames of the simultaneously recorded video shown above. The low-pass-filtered acceleration data is shown in dark blue, including a light blue version with a higher cut-off frequency.}
    \label{fig:perchcycle}
\end{figure*}

Following the static and free-fall tests, a series of free-flight tests were conducted to demonstrate the system's ability to perform a full perch cycle in a realistic environment. The generic perching cycle can hereby be divided into three phases, as clarified by figure~\ref{fig:perchcycle}. This cycle is initiated from the flying condition, with the landing phase, where the system approaches the perching object. This is followed, after successful attachment, by a perched phase, which enables the system to perform a defined mission or enter a state of rest. Lastly, the take-off phase is performed from a standstill, and the flying condition is again entered, completing the cycle.

During testing, our system's first phase (green strip in figure~\ref{fig:perchcycle}) begins in a generic free-flying condition. It consists of landing the drone on the perch through deceleration and a nearly vertical approach path, so that the gripping mechanism triggers upon contact and attaches the drone securely to the branch. In the second phase (blue strip), the servo-motor is automatically disconnected from power (500\,ms after impact) to preserve energy, and the drone is passively held atop the branch, with the motors shut off. This phase can be extended for as long as needed, depending on the mission. In the third phase (yellow strip), the drone's motors are rapidly engaged, and the take-off command is given in the form of an IR signal. This signal resets the ratchet mechanism and simultaneously opens the grippers to make take-off possible. Take-off is then typically performed near-vertically; however, a degree of deviation of up to 35° to the vertical is permitted, corresponding to the shape of the opened grippers.

The perching cycle was successfully performed several times in separate tests for landing and take-off and has also been demonstrated in continuous takes to show the viability of the system.
Video data was recorded both at 500\,fps, in slow motion, and at 60\,fps at 4K, with simultaneous IMU data recording for all tests.
Video footage of these flight tests can be found in Extension~\ref{Extensions}.3 for an exemplary slow-motion recording and in Extension~\ref{Extensions}.4 for the real-time recordings of the full perching cycle.

An example of a successful continuous perch-cycle can be seen in figure~\ref{fig:perchcycle}. The IMU Z-acceleration data is plotted against time-from-impact for the approach phase, and against time-from-release for the take-off phase. For compactness, the data for the perched state, where acceleration remains constant, is partly cut out. However, the data shown in the plot comes from a single recording, and the system stayed perched in a stable manner without power to any motors until the take-off sequence was initiated. In Extension~\ref{Extensions}.5, the full acceleration data of this uninterrupted perching cycle can be seen. To reduce the noise and obtain clearer results, the IMU data was filtered through a low-pass Butterworth filter. Two filters were considered, one with a cut-off frequency of 20\,Hz, still allowing for the flapping frequency of 19\,Hz to be observed, and the other with a lower cut-off frequency of 7\,Hz, thus excluding the flapping effects and presenting a clearer view. 

The perching sequence shown involved an impact speed of 0.82\,m/s with the characteristic impact peak in positive acceleration occurring directly after impact. Upon impact, the system oscillates for a short period of time, then settles while all motors are shut off and the stable perched state is reached. Following the longer perched phase, the take-off sequence is initiated around 300\ ms from release. This is visible in a sequence of initial oscillations in the acceleration connected to the wing motors being powered up shortly before release. After release, a phase of positive acceleration can be seen, leading to the system reaching an end velocity of 1.48\,m/s.

\begin{figure}
    \centering
    \includegraphics[width=0.9\columnwidth]{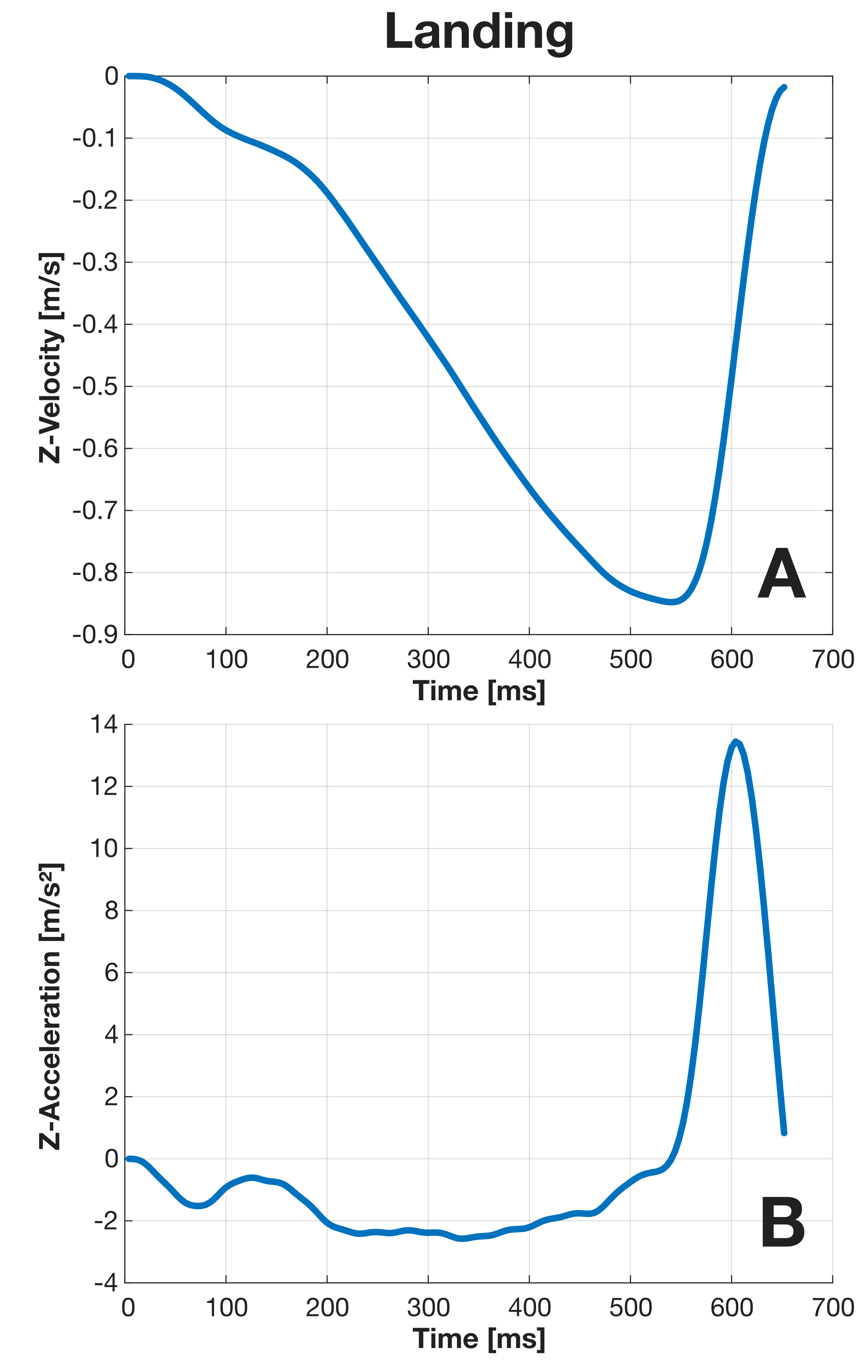}
    \caption{Detailed view on the acceleration and calculated velocity over time during a successful landing in the perching cycle, with data captured from the onboard IMU.}
    \label{fig:landing}
\end{figure}

\begin{figure}
    \centering
    \includegraphics[width=0.9\columnwidth]{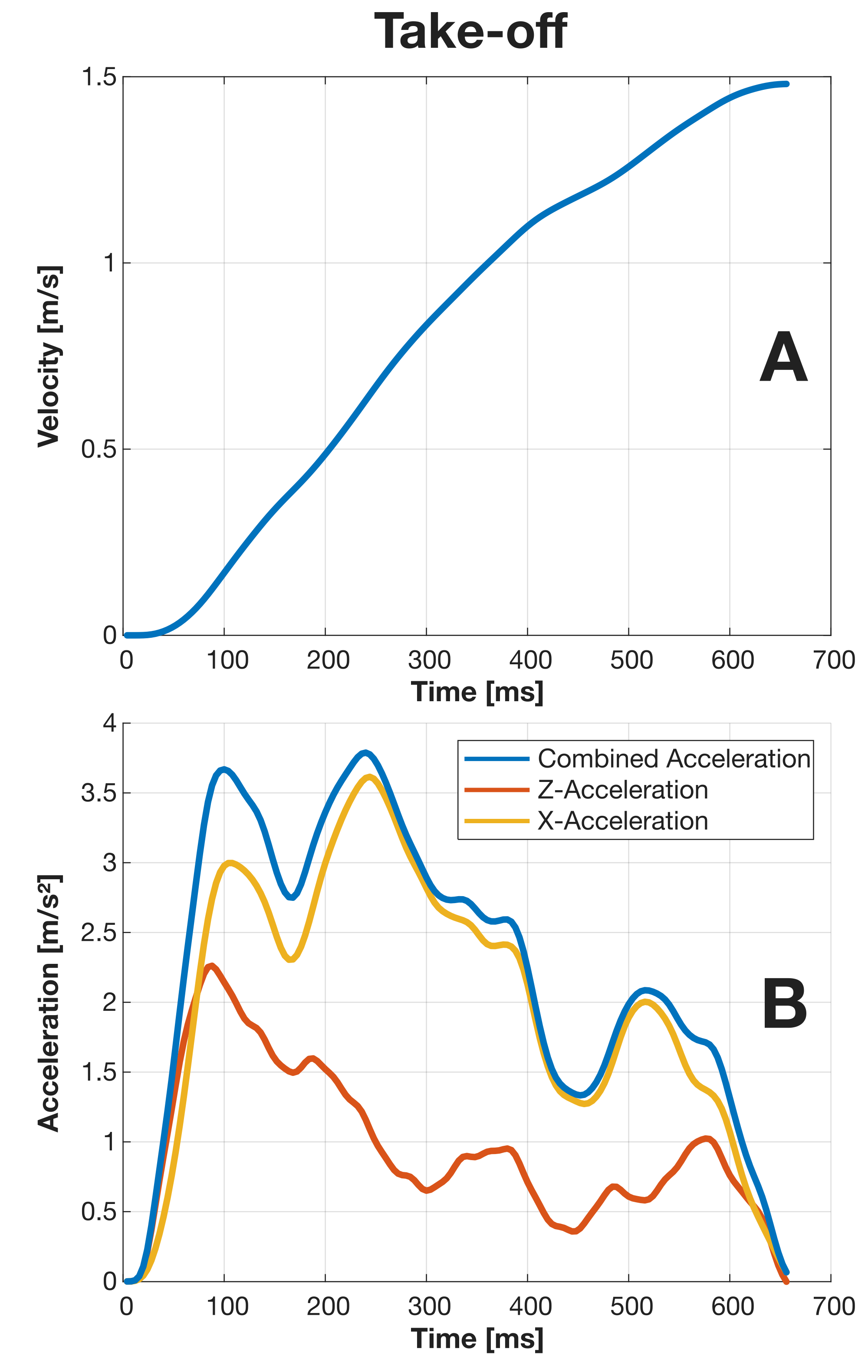}
    \caption{Detailed view on the acceleration and calculated velocity over time during take-off in the perching cycle, with data captured from the onboard IMU.}
    \label{fig:takeoff}
\end{figure}

A more detailed view of the collected data from a free-flight sequence during landing and subsequent take-off is shown in figure~\ref{fig:landing} and figure~\ref{fig:takeoff}, respectively.

Figure~\ref{fig:landing}.B shows the filtered Z-acceleration data over time, starting with the initiation of the landing sequence. Here, a large impact acceleration can be seen upon contact, where a more sophisticated trigger setup could be implemented to reduce this peak. However, the current design was able to withstand these forces without failure.
Figure~\ref{fig:landing}.A displays the integrated acceleration data, thus showing the velocity during approach and impact, all the way until standstill. Here, the steady approach and the deceleration time frame can be seen.

The take-off sequence in figure~\ref{fig:takeoff} begins at release time. As the flight trajectory is diagonal in the X-Z-plane (see coordinates in figure~\ref{fig:perchcycle}), the acceleration data is plotted in X and Z components in panel B. Panel A shows the integrated combined acceleration, i.e. the velocity during take-off until a continuous flight speed is reached. Due to the quick opening of the grippers upon receiving the release signal, the take-off sequence can be carried out with any desired start velocity, provided the flapping-wing MAV is given sufficient thrust to take flight.

The free-flight tests showed that the system enables a flapping-wing MAV of over 170\,g to land on a perch, attach to it in a stable manner, remain attached for an extended period of time, while conserving energy, and reliably take off when desired, thus completing a full perch cycle. While attaching a gripper to the FWMAV introduces a penalty on the flight time, the capability to "rest" on a perch or conduct a perched mission adds value and extends the flight time.
As the goal of our first prototype was to demonstrate the viability of the perching operation, a significant reduction in flight time was accepted. However, this is mostly due to the external control board and the unoptimised gripper prototype, which can be made lighter in future work.

During flight testing, a few strategies for achieving a successful perching attempt and subsequent take-off could be noted. In the approach, it was beneficial to position the MAV in a zone atop the perching object with the gripper axis and perching object axis in parallel. Then, in a powered descent, the alignment of these axes in a vertical plane was conducted, and upon achieving near coincidence, the throttle could be reduced to zero, letting the system enter a free-fall until contact with the perch. However, it should be noted that this approach strategy was helpful for manual steering and could be optimised and parallelised in an automated approach. Still, approximate axis alignment as a priority, followed by a rapid descent, would also be suitable for a simple automated system. For take-off, the timing of gripper release is of primary importance; here, the best strategy is to quickly engage the throttle, to reduce disturbances to the stability introduced by a slow flapping start, and upon achieving at least enough power to enter a hover state, the grippers can be safely opened again. In an automated system, this could be easily implemented and optimised timing-wise for a quick take-off.

While the emphasis of the current study was on demonstrating the full flight cycle, and thus testing was conducted by a pilot, the manoeuvre could potentially be automated, as mentioned, making the framework more functional and robust. Doing so would first require adapting the control system to handle the higher payload and the resulting change in flight behaviour. An analysis of the effects of an increased payload on the flight stability and performance of lightweight flapping-wing robots with perching capabilities can be found in \citet{vanBruegge2025}.

To demonstrate the abilities of the framework, a tailless flapping-wing MAV was used; however, integration into different types, e.g. tailed FWMAVs, could be feasible as well. To achieve this, the approach and take-off flight paths must be considered foremost to optimally orient the grippers, allowing a more horizontal approach path. Furthermore, a widened gripper opening angle could be introduced, allowing for safe take-off upon gripper release. Overall, the mechanism could be easily scaled in size and power, with the gripper component representing a further variable that can be adjusted depending on the desired use case and flapping-wing MAV size.

\section{Conclusion}
In this paper, we presented a framework for energy-efficient perching in mid-size tailless FWMAVs based on a bioinspired soft-gripping mechanism. The viability and performance of the framework were extensively tested in over two hundred perching tests of the integrated system, in manual tests, free-fall tests and controlled free flight. Furthermore, the complete functionality and full perch cycle of landing, continuous perching and take-off could be repeatedly demonstrated and recorded. The gripping mechanism itself weighs below 39\,g and was successfully integrated into an exemplary tailless FWMAV platform, the Flapper Nimble+, resulting in a system with roughly 170\,g take-off mass. The lightweight and soft grippers used for perching attach themselves to different types of objects without damaging the surface, while a custom trigger system ensures reliable and fast activation of the gripping process upon contact. A precise ratchet mechanism allows the robot to passively maintain a stable position on the perch without a power drain and enables take-off from the perch thanks to an integrated reset mechanism. With this framework, the full perch cycle of landing, perching and take-off can be repeated at will.
Future work will focus on developing control approaches to ensure safe and stable flight of lightweight flapping-wing robots with perching capabilities despite the considerable added payload of a gripping mechanism, as well as on achieving fully autonomous take-off and perched landing. Combining advanced control strategies with the proposed perching framework will significantly expand the versatility and capabilities of flapping-wing robots, taking them yet again a step closer to real-world applicability.

\appendix
\section*{Appendix}
\subsection{Index to Multimedia Extensions}

\begin{table*}[htb]
\caption{Index to Multimedia Extensions}
\label{Extensions}
\begin{tabular}{ccc}
\br
Extension & Media type & Description \\ \mr
1         & Data       & Velocity comparison of Vicon and IMU data during initial fall-testing \\
2         & Data       & Perch velocities at different inclinations during extended fall-testing \\
3         & Video      & Slow-motion video of landing and take-off       \\
4         & Video      & Full uninterrupted perching cycle               \\
5         & Data       & Filtered and unfiltered IMU acceleration data for the full uninterrupted perching cycle \\ \br
\end{tabular}
\end{table*}

\section*{Acknowledgements}
The authors received no financial support for the research, authorship, and/or publication of this article.

\bibliographystyle{unsrtnat}
\bibliography{references}

\end{document}